\documentclass[conference]{IEEEtran}
\IEEEoverridecommandlockouts
\usepackage{cite}
\usepackage{amsmath,amssymb,amsfonts}
\usepackage{algorithmic}
\usepackage[linesnumbered,lined,boxed,commentsnumbered]{algorithm2e}
\newcommand*{\Scale}[2][4]{\scalebox{#1}{$#2$}}%
\usepackage{titlesec}
\titlespacing*{\section}{0pt}{0.1\baselineskip}{0.2\baselineskip}
\usepackage{graphicx}
\usepackage{float}
\graphicspath{{images}}
\usepackage{subcaption}
\usepackage{hyperref}
\usepackage{textcomp}
\usepackage{xcolor}
\def\BibTeX{{\rm B\kern-.05em{\sc i\kern-.025em b}\kern-.08em
    T\kern-.1667em\lower.7ex\hbox{E}\kern-.125emX}}
\begin{document}
\title{Integrating Bidirectional Long Short-Term Memory with Subword Embedding for Authorship Attribution}

\makeatletter
\newcommand{\linebreakand}{%
  \end{@IEEEauthorhalign}
  \hfill\mbox{}\par
  \mbox{}\hfill\begin{@IEEEauthorhalign}
}
\author{\IEEEauthorblockN{1\textsuperscript{st} Abiodun Modupe}
\IEEEauthorblockA{\textit{Department of Computer Science} \\
\textit{University of Pretoria}\\
Pretoria, South Africa \\
abiodunmodupe@gmail.com}
\and
\IEEEauthorblockN{2\textsuperscript{nd} Turgay Celik}
\IEEEauthorblockA{\textit{School of Electrical \& Information Engineering} \\
\textit{University of the Witwatersrand}\\
Johannesburg, South Africa\\
celikturgay@gmail.com}
\and
\IEEEauthorblockN{3\textsuperscript{rd} Vukosi Marivate}
\IEEEauthorblockA{\textit{Department of Computer Science} \\
\textit{University of Pretoria}\\
Pretoria, South Africa \\
vukosi.marivate@cs.up.ac.za}
\linebreakand 
\IEEEauthorblockN{4\textsuperscript{th} Oludayo O. Olugbara}
\IEEEauthorblockA{\textit{MICT SETA Center of Excellence in 4IR} \\
\textit{Durban University of Technology}\\
 Durban, South Africa\\
oludayoo@dut.ac.za}
}
\maketitle
\begin{abstract}
The problem of unveiling the author of a given text document from multiple candidate authors is called authorship attribution. Manifold word-based stylistic markers have been successfully used in deep learning methods to deal with the intrinsic problem of authorship attribution. Unfortunately, the performance of word-based authorship attribution systems is limited by the vocabulary of the training corpus. Literature has recommended character-based stylistic markers as an alternative to overcome the hidden word problem. However, character-based methods often fail to capture the sequential relationship of words in texts which is a chasm for further improvement. The question addressed in this paper is whether it is possible to address the ambiguity of hidden words in text documents while preserving the sequential context of words. Consequently, a method based on bidirectional long short-term memory (BLSTM) with a 2-dimensional convolutional neural network (CNN) is proposed to capture sequential writing styles for authorship attribution. The BLSTM was used to obtain the sequential relationship among characteristics using subword information. The 2-dimensional CNN was applied to understand the local syntactical position of the style from unlabeled input text. The proposed method was experimentally evaluated against numerous state-of-the-art methods across the public corporal of CCAT50, IMDb62, Blog50, and Twitter50. Experimental results indicate accuracy improvement of 1.07\%, and 0.96\% on CCAT50 and Twitter, respectively, and produce comparable results on the remaining datasets.
\end{abstract}
\section{Introduction}\label{sec1}
The process of determining and extracting information about the author of anonymous online text snippets and documents is referred to as authorship attribution (AA).\cite{modupe2022post}. AA is useful in applications like digital forensics \cite{modupe2011exploring}, authorship verification (AV) \cite{nirkhi2016authorship}, author profiling (AP) \cite{fabien2020bertaa}, plagiarism detection (PD) \cite{foltynek2019academic}, and historical literature attribution \cite{neal2017surveying}. Many of these applications heavily depend on feature engineering and machine learning algorithms (MLAs) \cite{stamatatos2009survey,ding2017learning,vosoughi2016tweet2vec,wu2021exploring} to recognize writing styles and anonymous text authors without jeopardizing the confidentiality of the author \cite{altakrori2018arabic}.
However, most of the existing techniques apply quantitative methods such as word choice, punctuation, function words, and sentence structure \cite{stamatatos2009survey,wu2021exploring}, while other research studies incorporate content features such as part-of-speech (POS), character, and word n-gram to determine the author's writing style \cite{zheng2023review,wu2021exploring}. However, these methods are inadequate when miscreants mask their identities, reassemble the word order, disguise the IP address of the texts, or even use imagery text to circumvent the competence of text-mining methods  \cite{alonso2021writer,modupe2022post}.

The existing methods use conventional quantitative approaches such as vocabulary choice, punctuation, function words, and sentence structure \cite{stamatatos2009survey,wu2021exploring}, whereas other research studies incorporate content features such as part-of-speech (POS), character and word n-gram to characterize and recognize the writer's writing style \cite{zheng2023review,wu2021exploring}. However, these methods are inadequate when miscreants mask their identities, reassemble the word order, disguise the IP address of the texts \cite{alonso2021writer}, or even use the imagery in the text to undermine the usefulness of MLAs \cite{modupe2011exploring,modupe2014filtering}.

With the advancement of large language models (LLMs) such as Word2Vec and bidirectional encoder representations from transformers (BERT) \cite{liu2020character,devlin2018bert} that semantically map similar words to a common vector space with a similar vector representation to characterize inherent contextualization in a given text in a natural language, they have exhibited good performance. The methods are more parameterized, resulting in more gradual coverage and poor embedding vector representation. Additionally, finding the writing style that could be used to determine whether there is a single author or multiple authors is particularly challenging in a morphologically prosperous text with a long-tailed frequency distribution (such as online text snippets) \cite{liu2020character}.

In contrast to previous studies, in this paper, we introduce bidirectional long short-term memory (BLSTM) with a 2-dimensional (2D) convolution neural network (CNN) and a 2D max-pooling operation that uses byte-pair encoding (BPE) to transform given text into smaller units of word embedding. First and foremost, we make use of BLSTM layers to capture inherent semantic characteristics on both the time-step and sub-word feature dimensions. By leveraging the power of BLSTM layers, we are able to extract complex features that may not be immediately apparent to conventional methods of text classification. The use of the BPE algorithm further enhances the exploration of semantic representation by balancing character and word-level information, allowing for a more nuanced understanding of the language. This allows us to better understand the underlying patterns and relationships within the data, enabling more accurate predictions and classifications. Secondly, we feed the feature vectors into a 2D CNN and 2D max pooling to obtain more local syntactical information to represent the input text for AA tasks. By using vectors in a 2D CNN and 2D max pooling, the model is able to capture local features and patterns within the text. This hybrid approach is particularly helpful for AA tasks because it enables the model to pick up on subtle nuances and contextual cues that other methods might miss. This is particularly useful in natural language processing tasks, where understanding the nuances of language is crucial for success. With our approach, we succeed in achieving state-of-the-art (SOTA) outcomes on a range of benchmarks, demonstrating the effectiveness of our method. Overall, this approach represents a powerful tool for evaluating and understanding natural language in a variety of contexts.  The primary effects of this paper are summarized as follows:

\begin{itemize}
    \item The byte-pair-encoding (BPE) compression algorithm is used to represent sentences in a large corpus of data with a small set of subword units to deal with imaginary word difficulties and split the writing into characters, bytes, the frequency, and merge for the AA tasks. 
    \item We employ BLSTM to find an appropriate writing style that adequately characterizes an author's writing style through 2D convolutional and max-pooling techniques based on the BPE representation
    \item We add Gaussian noise to the network, in addition to BLSTM-2DCNN, to test the robustness of the proposed model, reduce the number of parameters in training, and improve performance on AA tasks.
    \item We perform experiments on three publicly available datasets (CCAT50, IMDb62, and Blog50) and a completely new Twitter dataset to exhibit that the proposed BLSTM-2DCNN model outperforms prior SOTA methods for the AA tasks.
\end{itemize}
The following wraps up the structure of the paper: In Section \ref{sec1}, we present preliminary information about the problem related to authorship attribution and the combination. In Section \ref{sec2}, we provide an overview of related work. Section \ref{sec3}, provides details about the structure of our proposed BLSTM-2D CNN max-pooling model for AA tasks. Section \ref{sec4}, detailed the dataset, parametric settings, and experimental results. Section \ref{sec5} concludes with the conclusions and future work. 
\section{Related Work}\label{sec2}
With an increasing number of studies using LLMs to extract stylistic features in analyzing textual data to uncover underlying author identity for advancing research in AA tasks, many depend on handcrafted characteristics, which are not generalizable, time-consuming, and labor-intensive \cite{ding2017learning,stamatatos2009survey,vosoughi2016tweet2vec}. For example, in \cite{bevendorff2019bias}, the author develops a basic and fairly flawed (BAFF) model for AV to figure out if two similar pieces of documents were composed by exactly the same writer. The key is to find a good representation using a character n-gram and trigram vector and then compute a style difference using a well-known distance measurement (e.g., TF-weighted, TF-IDF-weighted, and others). On the PAN15 dataset, the study achieves an accuracy of 0.73\%, an F1 of 0.69\%, and a precision of 0.82\%. \cite{plakias2008tensor} used tensors of the second order with 2500; most frequently 3-grams to represent stylistic components for a given text, and~\cite{muttenthaler2019authorship} hows the influence of punctuation marks with an \textit{n}-gram model for AA tasks while masking punctuation marks with an asterisk $(\ast)$ symbol. \cite{sapkota2015not}, the author used character-level n-gram features to emphasize the importance of sub-word features, such as suffixes and prefixes, for authorship-related tasks. \cite{plakias2008tensor} uses second-order tensor space models integrated with support vector machines (SVM) to find stylistic properties of texts through character n-grams with much fewer parameters consisting of the 2,500 most frequent 3-grams of the limited training data. The proposed representation was evaluated on the publicly available Reuter corpus volume 1 (RCV1) and achieved an accuracy of 67\% on CCAT50 and 81.40\% on the IMDB62 dataset. The author in \cite{koppel2011authorship} uses naive similarity-based methods along with a principal component analysis (PCA) to reduce numerous random features and determine the actual authors of a given textual document from an extremely large set of candidate authors. The study reached a precision of 93.2\% from a set of 10,000 blogs harvested in August 2004 from blogger.com, covering 42.2\% of the entire dataset with 1,000 authors. \cite{sari2018topic} evaluates four widely used datasets to investigate how different types of stylistic elements, such as lexical, syntactic, and combinations could improve the accuracy of the model to identify the authorship of a given textual document under different conditions. Also, fail to encode word order and syntactic features in the given textual documents due to sparse and high dimensional data representation \cite{liu2020character}. 

To address the problem of feature engineering for the AA task, \cite{ruder2016character} used a CNN model as well as character and word information to extract meaningful stylistic representations. They achieved an excellent result by testing the model on various datasets such as emails, blogs, and social media datasets such as tweets. However, the authors acknowledged that their dataset was based on celebrities, which includes terms such as hashtags and acronyms as features. \cite{zhang2018syntax} use aligned character-based CNN models with the word-based model to improve SOTA performance for AA systems. Despite the poor results, they used the data augmentation technique and demonstrated that the proposed models benefit from more training data than the original data and are better suited for real-world scenarios.  \cite{shrestha2017convolutional} used a similar model to the one in \cite{ruder2016character} and evaluated it by employing the same dataset as presented in \cite{schwartz2013authorship}, yielding good results for 50 candidate authors with an accuracy of 76\%. The limitation is that 30\% of the training set's data came from bots rather than humans, which makes changing the parametric values simple. \cite{sari2017continuous} implements n-gram features with the help of a neural network in combination with the SoftMax function and a multi-layer perceptron as a classification layer defined as a continuous model. The exploratory outcomes showed that the continuous model was capable of classifying the input documents with an accuracy of 70–75\%. In contrast to the author in \cite{fabien2020bertaa}, who employs a pre-trained BERT (called BertAA) with a dense layer and SoftMax function for AA tasks, BertAA was tested on email, blogs, and the IMDb62 dataset and achieved 97\% accuracy on the IMDb62 dataset and lower performance accuracy on the others. In this paper, we extend previous work \cite{modupe2022post} to extract lexical stylometric features using byte-pair encoding (BPE) and feed them into the BLSTM to capture syntactic features in the time-step dimension in forward and backward directions. Initially, we employ BPE to convert the input text into vectors and BLSTM to capture syntactic feature vector representation. And then the 2D-CNN and pooling operations are applied to capture a meaningful semantic feature vector representation that could learn an author's writing style from a given textual document.

\section{Proposed method}\label{sec3}
The first portion of the proposed method is presented in Fig.~\ref{fig1} adopts a byte-pair-encoding (BPE) algorithm in the text representation as an embedding tactic to transform the pure text into numerical representations. In the feature extraction as the second phase, we feed the embedding modules into a bidirectional LSTM to understand the underlying semantics and apply CNN max-pooling over time to capture the local spatial syntactical position on writing style from the input text. The classification phase consists of a fully connected layer and soft-max function, which is sufficient to fit the function which takes the features and outputs the classification result. Besides, we combined annealed Gaussian noise with training the model to learn the writing style representations for AA tasks, which helped avoid overfitting and achieved lower training loss. We evaluated the model's performance using \emph{k}-fold cross-validation with a large spectrum of public datasets for authorship attribution on CCAT50, IMDb62, Blog50, and the newly assembled Twitter dataset for the AA tasks.
\begin{figure*}
	\centering
	\includegraphics[width=\textwidth]{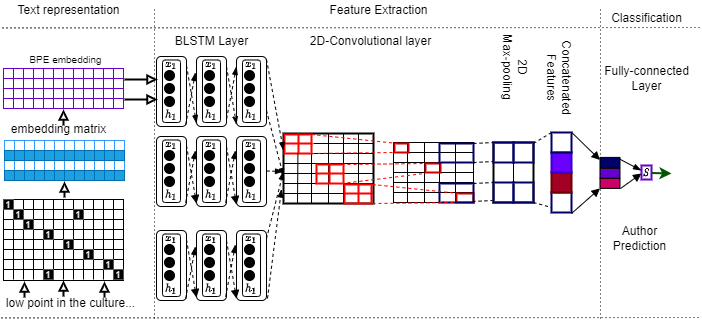}
	\caption{BLSTM-CNN model with subword information}
	\label{fig1}
\end{figure*}
\subsection{Text Representation}\label{3.1}
The first part of the proposed system is text representation based on character-level encoding that employed byte-pair encoding (BPE) to convert subwords in the pure text of the authors into numeric representations. The method brings prefers a balance between character and word-level hybrid representation to manage large-scale AA tasks.
\subsubsection{Sub-word embedding}\label{3.1.1}
BPE is a tokenization technique adopted in machine translation to deal with imaginary word problems or hidden writing in each text. It is unsupervised and requires no information about the author. The algorithm of BPE first initializes a symbol by splitting the input text into characters. Then, iteratively count all symbol pairs and replace each occurrence of the most frequent pair $\emph{(x,y)}$ with a new $\emph{xy}$ symbol and add it to the symbol set named ``merge operation''. Each merge operation generates a new symbol. The size of the final symbol set is equal to that of the first single character, plus the number of merge operations. The only hyperparameter of the BPE algorithm, as shown in Algorithm~\ref{alg1} is the number of merge operations and if the number of the merge operation is large, we will get a better vocabulary.
\setlength{\arraycolsep}{-3pt}
\begin{algorithm}
\caption{BPE algorithm \cite{vilar2021statistical}}\label{alg1}
\SetKwInOut{KwIn}{Input}
\SetKwInOut{KwOut}{Output}
\KwIn{training data $\Scale[0.8]D$ of words split into character sequence with number $\Scale[0.8]N$ of rules}
\KwOut{list $\Scale[0.8]K$ of $\Scale[0.8]N$ merge rules}
$\Scale[0.8]K:=[~]$\;
\lnl{InRes1}\While{\emph{length}~${(K)}~\leq~N$}
{
    $(x,y):=\underset{(x,y)}{\arg\max}\{\emph{count}_{D}{(x,y)}\}$\;
    $\textit{rule}:=\langle (x,y)\longrightarrow xy \rangle$\;
    $\Scale[0.8]D:=\emph{apply}(\textit{rule},\Scale[0.8]D)$\;
    $\Scale[0.8]K:=\emph{append}(\textit{rule},\Scale[0.8]K)$\;
}
\end{algorithm}

For untokenized unique text, we first split it into a single character and then iteratively do the merge operations following the merge order in the training step; until there are no more symbols that can be merged. If the number of merge operations is large, the token will tend to have more characters, and the granularity tends to be large. Otherwise, the granularity of the original text will be small. In our system, we do not use BPE as a compression algorithm. Instead, we use this algorithm to find sub-words as $\emph{n}$-grams with high frequencies for word segmentation, achieved if we joined characters together. However, we did not substitute them with new symbols. Table \ref{table1} shows an example of how subwords are obtained from a raw input text after $\mathcal{\Scale[0.8]N}$ iterations.

The text has now been subdivided into subword sequences. To use subword embedding to represent the text, we first create a one-hot vector for each subword type. The one-hot vector for the \textit{i}th subword in the vocabulary is a sparse binary vector $\emph{o}_{i}$ which has 1 as the \textit{i}th element and all 0 for others. After that, we project this embedding hyperspace onto a smaller hyperspace by multiplying the one-hot embedding with a subword embedding matrix $\mathcal{\Scale[0.8]S}$ with size $\mathcal{\Scale[0.8]N}\times\mathcal{\Scale[0.8]D}$, where $\mathcal{\Scale[0.8]N}$ represent the sub-word vocabulary size and $\mathcal{\Scale[0.8]D}$ is the dimension for the target embedding hyperspace. Therefore, we represent each sub-word information as a dense vector $s_i=\mathcal{\Scale[0.8]S}^{\Scale[0.8]T}_o{i}$, and the text with length $\mathcal{\Scale[0.8]T}$ is represented by a sequence of subword embedding vectors $(s_1,s_2,\cdots,s_{\Scale[0.8]T})$. Therefore, the subword embedding matrix $\mathcal{\Scale[0.8]S}$ are trained together.
\begin{table}[!t]
\renewcommand{\arraystretch}{1}
\caption{BPE example for input text.}
\label{table1}
\centering
\begin{tabular}{cl}
\hline
\textbf{Iteration} & \textbf{Words}\\
\hline
0 & workers work in workshop\\
1 & workersworkinworkshop\\
2 & \underline{\textit{wo}}rkers\underline{\textit{wo}}rkin\underline{\textit{wo}}rkinshop\\
3& \underline{\textit{wor}}~kers\underline{\textit{wor}}~kin\underline{\textit{wor}}~kshop\\
4& \underline{\textit{work}}~ers~\underline{\textit{work}}~in~\underline{\textit{work}}~shop\\
\textit{N}& \underline{\textit{work}}~\underline{\textit{er}}~s \underline{\textit{work}}~\underline{\textit{in}}~\underline{\textit{work}}~\underline{\textit{shop}}\\
\hline
\end{tabular}
\end{table}
\subsection{Feature Extraction}
The second part of our system is feature extraction based on sub-word embeddings, which feed into BLSTM and CNN modules and produce a neuron representing the probability of feature vectors belonging to a certain authorship. BLSTM uses the inherent grammatical relationships in the author's writing style to capture local syntactical information.
\subsubsection{Bidirectional LSTM Layer}
Long short-term memory (LSTM) was first proposed~\cite{schuster1997bidirectional} to overcome the gradient vanishing problem of recurrent neural networks (RNN). The idea is to introduce an adaptive gating mechanism, which decides the degree to keep the previous state and memorize the extracted features of the current input text. So, given a sequence of input text, $\Scale[0.5]X=\{x_1,x_2,\cdots,x_l\}$, where $l$ represents the length of the input text, LSTM processes the input as a piece of subword information. At each time-step $t$, the memory $c_t$ and the hidden state $\textit{h}_t$ are updated as following:
\setlength{\arraycolsep}{0.5pt}
\begin{equation}\label{eqn1}
\begin{split}\Scale[0.2]
f_t=\sigma(x_{t}\Scale[0.5]W_{x,f}+\textit{h}_{t-1}\Scale[0.5]W_{\textit{h},f}+b_f)\\
i_t=\sigma(x_{t}\Scale[0.5]W_{x,i}+\textit{h}_{t-1}\Scale[0.5]W_{\textit{h},i}+b_i)\\
 o_t=\sigma(x_{t}\Scale[0.5]W_{x,o}+\textit{h}_{t-1}\Scale[0.5]W_{\textit{h},o}+b_o)
\end{split}
\end{equation}

where $\Scale[0.5]W_{x,f}$ and $\Scale[0.5]W_{\textit{h},f}$ are the weights of the LSTM from input to forget gate and from the hidden state to the forget gate. The $\Scale[0.5]W_{x,i}$ and $\Scale[0.5]W_{\textit{h},i}$ are the weights from input to input gate and from the hidden state to the input gate. Similarly, $\Scale[0.5]W_{x,o}$ and $\Scale[0.5]W_{\textit{h},o}$ represent the weights from the input to the output gate and from the hidden state to the output gate. Finally, $\textit{b}_f, \textit{b}_i,\text{and},\textit{b}_o$ are the bias for the forget, input, and output gate. $\sigma$ denotes the logistic sigmoid function. The memory cell can be calculated as:
\setlength{\arraycolsep}{1pt}
\begin{equation}\label{eqn2}
    \hat{c}_{t}= \tanh({x_{t}\Scale[0.5]W_{x,c}+\textit{h}_{t-1}\Scale[0.5]W_{\textit{h},c}+b_c})
\end{equation}

where $\Scale[0.5]W_{x,c}$ and $\Scale[0.5]W_{h,c}$ represent the weights of the LSTM from input to the memory and from the hidden state to the memory, respectively, and the $\textit{b}_{c}$ denotes the bias. The memory cell at the time step $t$ was computed by
\setlength{\arraycolsep}{1pt}
\begin{equation}\label{eqn3}
    c_{t}= f_{t} \odot c_{t-1}+i_{t}\odot\hat{c}_{t}
\end{equation}

where $\otimes$ denoted element-wise multiplication. The hidden state can be updated as:
\setlength{\arraycolsep}{-1pt}
\begin{equation}\label{eqn4}
    \textit{h}_{t}= i_{t}\odot\tanh{(c_{t})} 
\end{equation}

The RNN model was forward and the output at the time steps $t$ depends on the past context as well as the hidden state, e.g. the future context. The author in \cite{schuster1997bidirectional} introduced a BLSTM to extend the unidirectional LSTM by introducing a second hidden layer, where the hidden connections flow in opposition temporal order. Therefore, the model is able to exploit information from both the past and the future. In this study, BLSTM is used to capture past and future writing style information. As shown in Fig. \ref{fig1}, the system contains two sub-networks for the forward (f) and backward (b) sequence context based on the subword embedding from the input text at each time step $t$ as follow:
\setlength{\arraycolsep}{-2pt}
\begin{equation}\label{eqn5}
\begin{split}
    \textit{h}^{f}_{t}= \sigma(x_{t}\Scale[0.5]W^{f}_{x,\textit{h}}+\textit{h}^{f}_{t-1}\Scale[0.5]W^{f}_{\textit{h},\textit{h}}+b^{f}_{\textit{h}})\\
    \textit{\textit{h}}^{b}_{t}= \sigma(x_{t}\Scale[0.5]W^{b}_{x,\textit{h}}+h^{b}_{t-1}\Scale[0.5]W^{b}_{\textit{h},\textit{h}}+b^{b}_{\textit{h}})
\end{split}
\end{equation}
The output at each time $t$ can be computed as:
\begin{equation}\label{eqn6}
     \textit{h}_{t}= [\textit{h}^{f}_{t} \otimes \textit{h}^{b}_{t}]\Scale[0.5]W_{\textit{h},o}+b_{o}
\end{equation}

where $\otimes$ is the element-wise sum to combine the forward and backward pass outputs.
\subsubsection{Convolutional Layer}
We can access the future and past contexts from the BLSTM layer, and $o_t$ is related to all other writing styles (or words) in the given text. In this study, we effectively treat the matrix as feature vectors, so $\Scale[0.8]1\Scale[0.8]D$ convolution and the max pooling operation were used to capture local syntactical information. For matrix $\Scale[0.8]H_t=h_1,h_2,\cdots,h_1$, $\Scale[0.8]H\in\mathbb{\Scale[0.8]R}^{l\times d^{w}}$ obtained from BLSTM layer, 
where $d^{w}$ is the size of the subword embeddings. Then narrow convolution is utilized to extract local  features information over $\Scale[0.8]H$. The convolution operation involves a filter $m\in\mathbb{\Scale[0.8]R}^{k\times d}$, which is applied to a window of $\emph{k}$ subwords and $d$ feature vectors. For instance, a feature $o_{i,j}$ is generated from a window of vectors $\Scale[0.8]H_\emph{i:i+k-1,j:j+d-1}$ as
\setlength{\arraycolsep}{-1pt}
\begin{equation}\label{eqn7}
     \emph{o}_{i,j}= \emph{f}(\emph{\textbf{m}}\cdot\emph{\Scale[0.5]H}_\emph{i:i+k-1,~j:j+d-1}+b )
\end{equation}

where $i$ ranges from $\Scale[0.8]1~\text{to}~{(\Scale[0.8]l-\Scale[0.8]{k+1})}$, $j$ ranges from $\Scale[0.8]1~\text{to}~(\Scale[0.8]d^{\Scale[0.8]{w}}-\Scale[0.8]{d+1})$, $(\cdot)$ represents dot product, $\Scale[0.8]b\in\mathbb{\Scale[0.8]R}$ is a bias and an $\emph{f}$ is a non-linear function similar to hyperbolic tangent. So, we applied the filter to each possible window of the BLSTM layer matrix $\emph{\Scale[0.8]H}$ to obtain a feature map $\mathbf{\Scale[0.8]O}$:

\setlength{\arraycolsep}{-1pt}
\begin{equation}\label{eqn8}
     \textit{\Scale[0.8]O}=[o_{1,1},o_{1,2},\cdots,o_{l-k+1,d^{w-d+1}}]
\end{equation}
\setlength{\arraycolsep}{0pt}

with $\textit{\Scale[0.8]O}\in\mathbb{\Scale[0.8]R}^{(l-k+1)\times(d^{w-d+1})}$ represent one convolution filter process. The convolution layer has multiple filters for the same size filter to learn complementary features or multiple kinds of filters with different sizes. Then 2D max pooling operation $p\in\mathbb{\Scale[0.8]R}^{(p_1\times p_2)}$ is utilized to obtain a fixed length vector by applying it to each possible window of matrix $\textit{O}$ to extract the maximum value:
\setlength{\arraycolsep}{-3pt}
\begin{equation}\label{eqn9} 
\textit{p}_{i,j}=\textit{\Scale[0.8]down}(\textit{\Scale[0.5]O}_{i:i+\textit{p}_{1},j:j+\textit{p}_{2}})
\end{equation}

where $\textit{\Scale[0.8]down}{(\cdot)}$~represents the map pooling operation function, $i=(1,1+\textit{p}_1,\cdots,1+(l-k+{1}/{\textit{p}_{1}-1})\cdot \textit{p}_1)$ and $j=(1,1+\textit{p}_2,\cdots,1+(d_{w}-d+{1}/{\textit{p}_{2}-1})\cdot \textit{p}_2)$. Then, we compute the pooling operation as follows:
\setlength{\arraycolsep}{-2pt}
\begin{equation}\label{eqn10} 
\begin{split}
    \textit{h}^{\ast}&=[\textit{p}_{1,1},\textit{p}_{1},_{1+\textit{p}_{2}},\cdots,\textit{p}_{1}+ (l-k+1/\textit{p}_{1}-1)\cdot\textit{p}_{1},\\
    & 1+(d_{w}-d+1/\textit{p}_{2}-1)\cdot\textit{p}_{2}]
\end{split}
\end{equation}

where $\textit{h}^{\ast}\in\mathbb{\Scale[0.8]R}$, and the length of $\textit{h}^{\ast}$ is $\lfloor l-k+1/\textit{p}_1\rfloor\times\lfloor d_{w}-d+{1}/{\textit{p}_{2}-1} \rfloor$.

\subsubsection{Classification Layer}
For the AA tasks, the output $\textit{h}^{\ast}$ from the max pooling was passed over the fully-connected layer of the input text $\Scale[0.8]X$, then feeds it to a softmax function as a classifier to predict the inherent writing style related to a particular candidate author $\hat{\textit{y}}$ from the set of a discrete set of author $\textit{\Scale[0.8]Y}$. So, the classifier takes the hidden state $\textit{h}^{\ast}$ as input as follows:
\setlength{\arraycolsep}{-3pt}
\begin{equation}\label{eqn11}
         \hat{\textit{p}}(\textit{y}|\textit{x})=\emph{softmax}\big(\Scale[0.8]W^{(\textit{x})}\textit{h}^{\ast} + \textit{b}^{(\textit{x})}\big)
\end{equation}
\setlength{\arraycolsep}{-3pt}
\begin{equation}\label{eqn12}
  \hat{\textit{y}}=\operatorname*{argmax}_{y\to\infty}\hat{p}(y|x)
\end{equation}

To learn the model parameters, we minimize the cross-entropy loss as the training objective by calculating the loss as a regularized sum:
\setlength{\arraycolsep}{-3pt}
\begin{equation}\label{eqn13}
J{(\theta)}=-\frac{1}{m}\sum_{i=1}^{m}t_{i}\log(y_i)+\lambda||\theta||^{2}_{F}
\end{equation}

where $\textbf{\textit{t}}\in\mathbb{\Scale[0.8]R}^{m}$ denote the one-hot encoding for the ground truth values, $\textbf{\textit{y}}\in\mathbb{\Scale[0.8]R}^{m}$ is the predicted probability of the candidate author by softmax, $m$ is the number of expected target authors, and $\lambda$ is the $\ell_2$ regularization parameter. Training is done through the Adam Optimization algorithm~\cite{kingad2015methodforstochasticoptimization} as further explained in Section \ref{sec4.2}. Finally, the pseudocode of our model is given in Algorithm \ref{alg2}, where we use simplified variables to make the procedure clear.

\setlength{\arraycolsep}{-3pt}
\begin{algorithm*}[ht]
\caption{Pseudocode for BLSTM-2DCNN max-pooling with subword information}
\label{alg2}
\SetKwInOut{KwIn}{Input}
\SetKwInOut{KwOut}{Output}
\KwIn{Training data $X=\{X_i\}_{i=1}^{n}$~ \emph{m} is the batch size,~\emph{n} is the training samples, \emph{w} is the parameters and \emph{l} is the length of the input text.}
\KwOut{trained model.}
Initialize $w$ randomly\;
\ForEach{iteration}{
    \ForAll{$k\in\{1,2,\cdots, \lceil \frac{n}{m} \rceil \}$}{
    Sample from each batch $\textbf{\Scale[0.8]X}_{\textit{k}}~\text{from}~\textbf{\Scale[0.8]X}$\;
    Divide each sample in the batch into the sequence $\{\textbf{\Scale[0.8]X}_{\textit{k}}^{1},\textbf{\Scale[0.8]X}_{\textit{k}}^{2},\cdots,\textbf{\Scale[0.8]X}_{\textit{k}}^{l}\}$\;
    Feed sequential batch into BLSTM consisting of forward and backward neuron, respectively, and outcome two output sequence $\{ \textit{\textbf{h}}_{fk}^{1},\textit{\textbf{h}}_{fk}^{2},\cdots,\textit{\textbf{h}}_{fk}^{l}, \textit{\textbf{h}}_{bk}^{1},\textit{\textbf{h}}_{bk}^{2},\cdots,\textit{\textbf{h}}_{bk}^{l} \}$~~ Equation~\ref{eqn5}\;
    Concatenate the BLSTM layer to obtain $\textit{\Scale[0.8]H}\in\mathbb{\Scale[0.8]R}^{l\times d^{w}}$, the narrow convolution is used to extract local dependent features over $\textit{H}$ to produce a feature map $\textit{\Scale[0.8]O}$  \;
    Then, $p\in\mathbb{\Scale[0.8]R}^{(p_1\times p_2)}$ is applied to each possible window of matrix $\textit{\Scale[0.8]O}$ to obtain $\textit{h}^{\ast}$~~Equation~\ref{eqn10} represent the stylometric representation (e.g., writing styles) of the input $\textbf{\Scale[0.8]X}$\;
    Feed the output $\textit{h}^{\ast}$ into the softmax classifier layer and obtain the classification result in Equation~~\ref{eqn11}~and \ref{eqn12} respectively\;
    Update $\textit{w}$ by minimizing with the cross-entropy loss in Equation~\ref{eqn13} using Adman algorithm \cite{kingad2015methodforstochasticoptimization}\;
    }
}
\end{algorithm*}
\setlength{\arraycolsep}{2pt}
\section{Experiment and results}\label{sec4}
\subsection{Datasets}
We tested our model on CCAT50, IMDb62, Blog50, and the new Twitter datasets, all of which were available to the public and had a wide range of authorship information. These datasets have different numbers of authors and sizes of documents, which lets us run tests and see how our methods work in different situations. The descriptive statistics for the datasets are shown in Table~\ref{table2}, where ~\textit{n} represents the number of authors, \textit{w} represents the average number of tokens per text, \textit{c} represents the average number of words per and \textit{t} is the number of documents. CCAT50 has a total of 5,000 documents written by 50 authors~\cite{stamatatos2011plagiarism}. The Internet Movie Database (IMDB) contains 62,000 movie reviews and 17,550 message posts from 62 prolific authors~\cite{seroussi2014authorship}. The Blog50 original contains 681,288 posts by 19,320 bloggers, and we select posts written by the top 50 bloggers~\cite{schler2006effects}. For the Twitter dataset,~\cite{rocha2016authorship} had already assembled a dataset for AA tasks at the beginning of 2019; we proceeded with the collection of new and bigger ones as a means to have up-to-date information and to analyze more recent scenarios, where the limitation of characters has increased from 140 to 280 characters that Twitter has just released. We created the dataset using the Twitter API to aggregate a list of 4,391 celebrities such as columnists, politicians, technologists, artists, and influencers on social media in 68 categories covering everything from politics, social unrest, and tech to arts and culture. We gathered about 130 million tweets from over 15,000 authors, chosen at random from a Twitter API search for stop-words. For our experiment, each dataset is split by sampling 60\% of each author's documents into a training set, 20\% for validation, and the remainder for testing over 10-fold cross-validation as used in most AA tasks.
\begin{table}[!t]
\renewcommand{\arraystretch}{1.0}
 \caption{Statistics of the datasets.}
\label{table2} 
\centering
    \begin{tabular}{lcccc}
        \hline
        Datasets&\textit{n}&\textit{w}&\textit{c}&\textit{t}\\
\hline
       CCAT50&50&584&3,010&5,000\\
       IMDb62&62&345&1,742&62,985\\
       Blog50&50&117&542 &681,288\\
       Twitter50&50&36&119&130,109,964\\ 
       \hline
    \end{tabular}
\end{table}
\subsection{Hyperparameters}\label{sec4.2}
For all the datasets, we used Adam optimization \cite{kingad2015methodforstochasticoptimization} with a mini-batch size of 64, a ReduceLROnPlateau~\cite{Gabruseva_2020_CVPR_Workshops} with the patience of 5, and learning rate of 0.0001 with 5 epochs and a decrease factor of 0.5. During training, the dimension of the subword vector is 300. The hidden units of BLSTM are 128. We use 100 convolutional filters for the window sizes of (3,3) with a max-pooling size of (2,2). For regularization, we employ dropout operation with a dropout rate of 0.5 for the subword embeddings, 0.3 for the BLSTM, and 0.2 for the penultimate layer with Gaussian Noise of 0.2 active at the training time. We also use the $\ell_2$ penalty with coefficient $10^{-5}$ over the parameter and trained for 20 epochs. All word vectors and feature vectors are randomly initialized and learned, and updated during the training process. The dimensions of the word vector and hidden layer size are d=64 in all models. We use 128 convolutional filters, each for window sizes of (3,3) and 2D pooling sizes of (2,2). All the experiments were repeated 5-fold cross-validation, and the accuracy reported represents the mean average of the prediction writing style of the authors.
\subsection{Results}
The overall authorship attribution accuracies of our methods and the baseline are provided in Table~\ref{table3}. The missing values denoted as ``(--)'' in Table~\ref{table3} indicate that the feature and the model were not considered by the previous SOTA methods. This study implements three models, BLSTM-2DCNN, BLSTM-2DCNN word embedding, and BLSTM-2DCNN gradient noise with subword information. Table~\ref{table3} presents the performance of all three models and other state-of-the-art models on four datasets for authorship-based tasks.
\begin{table}[!t]
\centering
\caption{Performance comparison accuracy.}\label{table3}
\setlength{\tabcolsep}{1pt}
\begin{tabular}{p{4cm}cccc}
\hline
   \textbf{Models} & \textbf{CCAT50}& \textbf{IMDb62}& \textbf{Blog50}& \textbf{Twitter50}\\
\hline
SVM with 3-gram~\cite{plakias2008tensor} & 67.00 & 81.40 & -- & --\\
Imposters~\cite{koppel2011authorship} & -- & 76.90 & 26.00 & 52.50\\
LDAH-S with topics~\cite{seroussi2011authorship} & -- & 72.00 & 18.30 & 38.30\\
SVM Affix + punctuation~\cite{sapkota2015not} & 69.30 & 69.90 & -- & --\\
Style, content \& hybrid~\cite{sari2018topic} & -- & 75.76 & 84.51 & --\\
\hline
CNN-character~\cite{ruder2016character} & -- & 91.70 & 49.40 & 86.80\\
CNN-word~\cite{ruder2016character} & -- & 84.30 & 43.00 & 80.50\\
CNN n-gram~\cite{shrestha2017convolutional} & 76.50 & 95.21 & 53.09 & --\\
Continuous n-gram~\cite{sari2017continuous} & 72.60 & 95.12 & 52.82 & --\\
Syntax-CNN~\cite{zhang2018syntax} & 81.00 & \textbf{96.16} & 56.73 & --\\
\hline
LSTM word embedding~\cite{gupta2017study}  & 61.47 & -- & -- & --\\
GRU word embedding~\cite{gupta2017study}  & 69.20 & -- & -- & --\\
Style-HAN~\cite{jafariakinabad2019style} & 82.35 & -- & \textbf{61.19} & --\\
BertAA~\cite{fabien2020bertaa} & -- & 90.70 & 59.70 & --\\
\hline
BLSTM-2DCNN & 73.25 & 81.25 & 48.15 & 49.52\\
BLSTM-2DCNN WE\footnote{word embedding} & 74.50 & 89.72 & 53.26 & 79.30\\
BLSTM-2DCNN SG\footnote{subword+Gaussian noise} & \textbf{83.42} & 93.72 & 60.67 & \textbf{87.76}\\
\hline
\end{tabular}
\end{table}
As shown in Table~\ref{table3}, the BLSTM-2DCNN+Gaussian noise with subword embedding achieves comparative performance on three out of four datasets. Gaussian noise was combined with $\ell_{2}$ regularization to gain roughly 10\% better performance when compared to both traditional methods and the existing CNN-based models. Essentially, it achieves 2.9\%, 1.6\%, and 0.9\% test accuracy on CCAT50, Blog50, and Twitter datasets, respectively. In addition, the performance of the proposed model is superior to that of the CNN and BertAA model~\cite{ruder2016character,fabien2020bertaa}, which shows that learning from characters or leveraging on the pre-trained language model without feature engineering tasks can help to improve the performance for AA tasks. Our method is much better than the BertAA model, which validates the effectiveness of integrating a pre-trained BERT~\cite{devlin2018bert} language model with an extra dense layer to perform authorship classification. In addition, different from existing CNN-based methods, we leveraged the extracted features employing the BPE algorithm to represent words by their index in the vocabulary together with its subword vector classes. Consequently, the proposed model inherits the advantage of both traditional CNN-LSTM model~\cite{ruder2016character,gupta2017study,jafariakinabad2019syntactic} and BPE algorithms with Gaussian noise, which contributes to the performance improvement for the AA dataset as shown in Table~\ref{table3}.

The proposed algorithm outperforms all the CNN variants by more than 10\% and 4\% for both Blog and Twitter datasets with 50 authors. Differences for the IMDb62 domain render less discriminatory words or character sequences when authors review similar movies. The BertAA model is boosted on IMDb62 because they are less sensitive to topical divergence. However, they are less helpful in short-digit text like Blog50 and Twitter domain, where hashtags or emoticons are the most characteristic features.

To further substantiate the effectiveness of our model, we tested CNN and BertAA models and the Gaussian noise, respectively. We then reported the performance of the results in Table~\ref{table4}. For the CNN-based and BertAA model, we add Gaussian noise before the softmax classifier on the same network structure. Comparing CNN-based and BertAA models, we see that each model can improve authorship classification accuracy using the same extracted features from BPE algorithms. In addition, it can be seen in Table~\ref{table4} that our model is superior to the counterpart CNN-based or BertAA model with a pre-trained weighted vector. Fig.~\ref{fig2} depicted the accuracy and convergence curves for all datasets, while Fig.~\ref{fig3}  represented the best accuracy with the fastest fast convergence between the proposed model and the CNN-char model~\cite{ruder2016character} on the Twitter dataset.

\begin{table}[!t]
\centering
\caption{Other performance}\label{table4}
\setlength{\tabcolsep}{1.5pt}
\begin{tabular}{lcccc}
\hline
\textbf{Models} & \textbf{CCAT50}& \textbf{IMDb62}& \textbf{Blog50}& \textbf{Twitter50}\\
\hline
CNN-char~\cite{ruder2016character} & 76.77 & 91.01 & 59.70 & 84.29\\
CNN-word~\cite{ruder2016character} & 77.03 & 92.30 & 62.40 & 82.50\\
BertAA~\cite{fabien2020bertaa} & 78.90 & 93.00 & 64.40 & 62.50\\
\hline
Our method&83.42 & 93.72 & 60.67 & 87.76\\
\hline
\end{tabular}
\end{table}

\begin{figure}[H]
    \centering
    \begin{subfigure}{0.45\linewidth}
        \includegraphics[scale=0.3]{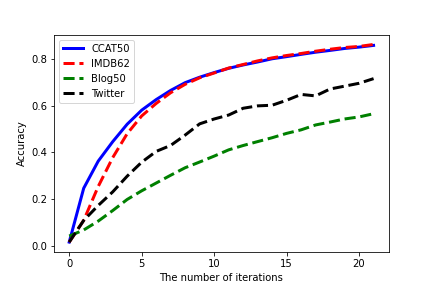}\label{a}
    \end{subfigure}
    \hskip1em
    \begin{subfigure}{0.45\linewidth}
        \includegraphics[scale=0.3]{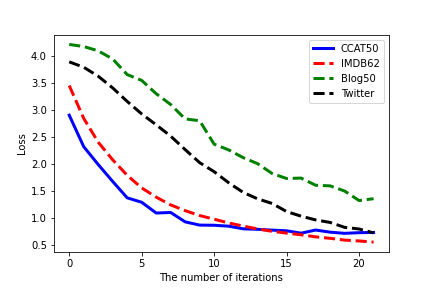}\label{b}
    \end{subfigure}
    \caption{The accuracy and convergence curve over iterations on all the datasets}\label{fig2}
\end{figure}
\begin{figure}[H]
    \centering
    \begin{subfigure}{0.45\linewidth}
        \includegraphics[scale=0.3]{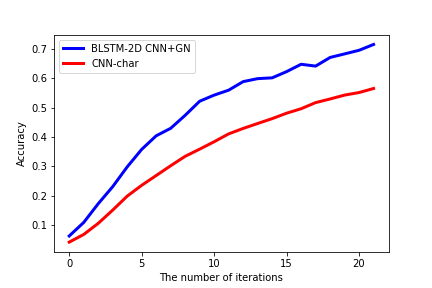}\label{a}
    \end{subfigure}
    \hskip1em
    \begin{subfigure}{0.45\linewidth}
        \includegraphics[scale=0.3]{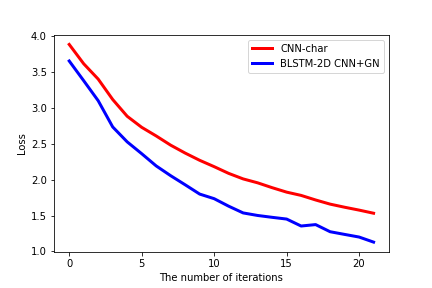}\label{b}
    \end{subfigure}
    \caption{The accuracy and convergence curve over iterations on Twitters dataset}\label{fig3}
\end{figure}
\subsection{Discussion}
The proposed BLSTM-2DCNN model outperforms most investigated methods by 1.07\% and 0.96\% on the CCAT50 and Twitter datasets, but is less superior on the IMDB62 and Blog50 datasets by 2.44\% and 0.32\%, respectively. The Style-HAN method, which encodes syntactic patterns using only POS tags, performs well on the CCAT50 dataset but has negligible results with Blogs datasets. CNN-word and CNN-word-word achieve worse performance in short-message domains like blogs and Twitter but provide better performance in domains where topical information is not discriminatory. The proposed BLSTM-2DCNN model is superior in capturing better stylistic information, such as words, topics, punctuation, or phrases re-posted by some authors when publishing comments. The shallow neural network \cite{sari2017continuous} beats the SVM \cite{plakias2008tensor,sapkota2015not} with similar feature sets, while the proposed BLSTM-2DCNN model shows superior encoding ability for AA tasks on all datasets. The BLSTM-2DCNN model outperforms SVM with frequent 3-grams, LDAH-S, Imposters, CNN variants~\cite{ruder2016character,zhang2018syntax,fabien2020bertaa}, and other deep neural network methods on 2 out of 4 datasets, except for the CCAT50 and the new Twitter dataset.

\section{Conclusion}\label{sec5}
This paper demonstrates that input-embedded vectors employing subword information feed with the BLSTM-2DCNN model could learn stylometric representations of different linguistic modalities for AA tasks. It showcases such a con- figuration’s effectiveness in dealing with common spelling errors from unstructured texts due to orthography and phonetic reasons, then learns stylistic and topical information to classify the author. In addition, the Gaussian noise is introduced to the fully conventional layers, which substantially reduces the large number of parameters arising from the model structure. Thus, the convergence rate of the model significantly speeds up and improves the classification accuracy. We evaluated the model against the state-of-the-art methods for an extensive range of authors, demonstrating the proposed model’s effectiveness in handling morphological variance and its applicability across authorship-related tasks. Future works will explore combining the model with a self-attention mechanism to model different linguistic levels such as structure, POS tagging, dependency, and semantics, applying subword information to improve the alignment of words in the input texts during training to advance the research on authorship-based tasks. 
\bibliographystyle{splncs04}
\bibliography{references.bib}

\begin{thebibliography}{10}
\providecommand{\url}[1]{\texttt{#1}}
\providecommand{\urlprefix}{URL }
\providecommand{\doi}[1]{https://doi.org/#1}

\bibitem{alonso2021writer}
Alonso-Fernandez, F., Belvisi, N.M.S., Hernandez-Diaz, K., Muhammad, N., Bigun,
  J.: Writer identification using microblogging texts for social media
  forensics. IEEE Transactions on Biometrics, Behavior, and Identity Science
  (2021)

\bibitem{altakrori2018arabic}
Altakrori, M.H., Iqbal, F., Fung, B.C., Ding, S.H., Tubaishat, A.: Arabic
  authorship attribution: An extensive study on twitter posts. ACM Transactions
  on Asian and Low-Resource Language Information Processing (TALLIP)
  \textbf{18}(1),  1--51 (2018)

\bibitem{bevendorff2019bias}
Bevendorff, J., Hagen, M., Stein, B., Potthast, M.: Bias analysis and
  mitigation in the evaluation of authorship verification. In: Proceedings of
  the 57th Annual Meeting of the Association for Computational Linguistics. pp.
  6301--6306 (2019)

\bibitem{devlin2018bert}
Devlin, J., Chang, M.W., Lee, K., Toutanova, K.: Bert: Pre-training of deep
  bidirectional transformers for language understanding. arXiv preprint
  arXiv:1810.04805  (2018)

\bibitem{ding2017learning}
Ding, S.H., Fung, B.C., Iqbal, F., Cheung, W.K.: Learning stylometric
  representations for authorship analysis. IEEE transactions on cybernetics
  \textbf{49}(1),  107--121 (2017)

\bibitem{fabien2020bertaa}
Fabien, M., Villatoro-Tello, E., Motlicek, P., Parida, S.: Bertaa: Bert
  fine-tuning for authorship attribution. In: Proceedings of the 17th
  International Conference on Natural Language Processing (ICON). pp. 127--137
  (2020)

\bibitem{foltynek2019academic}
Folt{\`y}nek, T., Meuschke, N., Gipp, B.: Academic plagiarism detection: a
  systematic literature review. ACM Computing Surveys (CSUR)  \textbf{52}(6),
  1--42 (2019)

\bibitem{Gabruseva_2020_CVPR_Workshops}
Gabruseva, T., Poplavskiy, D., Kalinin, A.: Deep learning for automatic
  pneumonia detection. In: Proceedings of the IEEE/CVF Conference on Computer
  Vision and Pattern Recognition (CVPR) Workshops (June 2020)

\bibitem{gupta2017study}
Gupta, B., Negi, M., Vishwakarma, K., Rawat, G., Badhani, P., Tech, B.: Study
  of twitter sentiment analysis using machine learning algorithms on python.
  International Journal of Computer Applications  \textbf{165}(9),  29--34
  (2017)

\bibitem{jafariakinabad2019style}
Jafariakinabad, F., Hua, K.A.: Style-aware neural model with application in
  authorship attribution. In: 2019 18th IEEE International Conference On
  Machine Learning And Applications (ICMLA). pp. 325--328. IEEE (2019)

\bibitem{jafariakinabad2019syntactic}
Jafariakinabad, F., Tarnpradab, S., Hua, K.A.: Syntactic recurrent neural
  network for authorship attribution. arXiv preprint arXiv:1902.09723  (2019)

\bibitem{kingad2015methodforstochasticoptimization}
KingaD, A.: A methodforstochasticoptimization. Anon. InternationalConferenceon
  Learning Representations. SanDego: ICLR  (2015)

\bibitem{koppel2011authorship}
Koppel, M., Schler, J., Argamon, S.: Authorship attribution in the wild.
  Language Resources and Evaluation  \textbf{45}(1),  83--94 (2011)

\bibitem{liu2020character}
Liu, B., Zhou, Y., Sun, W.: Character-level text classification via
  convolutional neural network and gated recurrent unit. International Journal
  of Machine Learning and Cybernetics  \textbf{11},  1939--1949 (2020)

\bibitem{modupe2022post}
Modupe, A., Celik, T., Marivate, V., Olugbara, O.O.: Post-authorship
  attribution using regularized deep neural network. Applied Sciences
  \textbf{12}(15), ~7518 (2022)

\bibitem{modupe2011exploring}
Modupe, A., Olugbara, O.O., Ojo, S.O.: Exploring support vector machines and
  random forests to detect advanced fee fraud activities on internet. In: 2011
  IEEE 11th International Conference on Data Mining Workshops. pp. 331--335.
  IEEE (2011)

\bibitem{modupe2014filtering}
Modupe, A., Olugbara, O.O., Ojo, S.O.: Filtering of mobile short messaging
  service communication using latent dirichlet allocation with social network
  analysis. In: Transactions on Engineering Technologies, pp. 671--686.
  Springer (2014)

\bibitem{muttenthaler2019authorship}
Muttenthaler, L., Lucas, G., Amann, J.: Authorship attribution in fan-fictional
  texts given variable length character and word n-grams. In: CLEF (Working
  Notes) (2019)

\bibitem{neal2017surveying}
Neal, T., Sundararajan, K., Fatima, A., Yan, Y., Xiang, Y., Woodard, D.:
  Surveying stylometry techniques and applications. ACM Computing Surveys
  (CSUR)  \textbf{50}(6),  1--36 (2017)

\bibitem{nirkhi2016authorship}
Nirkhi, S., Dharaskar, R., Thakare, V.: Authorship verification of online
  messages for forensic investigation. Procedia Computer Science  \textbf{78},
  640--645 (2016)

\bibitem{plakias2008tensor}
Plakias, S., Stamatatos, E.: Tensor space models for authorship identification.
  In: Hellenic Conference on Artificial Intelligence. pp. 239--249. Springer
  (2008)

\bibitem{rocha2016authorship}
Rocha, A., Scheirer, W.J., Forstall, C.W., Cavalcante, T., Theophilo, A., Shen,
  B., Carvalho, A.R., Stamatatos, E.: Authorship attribution for social media
  forensics. IEEE transactions on information forensics and security
  \textbf{12}(1),  5--33 (2016)

\bibitem{ruder2016character}
Ruder, S., Ghaffari, P., Breslin, J.G.: Character-level and multi-channel
  convolutional neural networks for large-scale authorship attribution. arXiv
  preprint arXiv:1609.06686  (2016)

\bibitem{sapkota2015not}
Sapkota, U., Bethard, S., Montes, M., Solorio, T.: Not all character n-grams
  are created equal: A study in authorship attribution. In: Proceedings of the
  2015 conference of the North American chapter of the association for
  computational linguistics: Human language technologies. pp. 93--102 (2015)

\bibitem{sari2018topic}
Sari, Y., Stevenson, M., Vlachos, A.: Topic or style? exploring the most useful
  features for authorship attribution. In: Proceedings of the 27th
  International Conference on Computational Linguistics. pp. 343--353 (2018)

\bibitem{sari2017continuous}
Sari, Y., Vlachos, A., Stevenson, M.: Continuous n-gram representations for
  authorship attribution. In: Proceedings of the 15th Conference of the
  European Chapter of the Association for Computational Linguistics: Volume 2,
  Short Papers. pp. 267--273 (2017)

\bibitem{schler2006effects}
Schler, J., Koppel, M., Argamon, S., Pennebaker, J.W.: Effects of age and
  gender on blogging. In: AAAI spring symposium: Computational approaches to
  analyzing weblogs. vol.~6, pp. 199--205 (2006)

\bibitem{schuster1997bidirectional}
Schuster, M., Paliwal, K.K.: Bidirectional recurrent neural networks. IEEE
  transactions on Signal Processing  \textbf{45}(11),  2673--2681 (1997)

\bibitem{schwartz2013authorship}
Schwartz, R., Tsur, O., Rappoport, A., Koppel, M.: Authorship attribution of
  micro-messages. In: Proceedings of the 2013 Conference on Empirical Methods
  in Natural Language Processing. pp. 1880--1891 (2013)

\bibitem{seroussi2011authorship}
Seroussi, Y., Zukerman, I., Bohnert, F.: Authorship attribution with latent
  dirichlet allocation. In: Proceedings of the fifteenth conference on
  computational natural language learning. pp. 181--189 (2011)

\bibitem{seroussi2014authorship}
Seroussi, Y., Zukerman, I., Bohnert, F.: Authorship attribution with topic
  models. Computational Linguistics  \textbf{40}(2),  269--310 (2014)

\bibitem{shrestha2017convolutional}
Shrestha, P., Sierra, S., Gonz{\'a}lez, F.A., Montes, M., Rosso, P., Solorio,
  T.: Convolutional neural networks for authorship attribution of short texts.
  In: Proceedings of the 15th Conference of the European Chapter of the
  Association for Computational Linguistics: Volume 2, Short Papers. pp.
  669--674 (2017)

\bibitem{stamatatos2009survey}
Stamatatos, E.: A survey of modern authorship attribution methods. Journal of
  the American Society for information Science and Technology  \textbf{60}(3),
  538--556 (2009)

\bibitem{stamatatos2011plagiarism}
Stamatatos, E., Koppel, M.: Plagiarism and authorship analysis: introduction to
  the special issue. Language Resources and Evaluation  \textbf{45}(1), ~1--4
  (2011)

\bibitem{vilar2021statistical}
Vilar, D., Federico, M.: A statistical extension of byte-pair encoding. In:
  Proceedings of the 18th International Conference on Spoken Language
  Translation (IWSLT 2021). pp. 263--275 (2021)

\bibitem{vosoughi2016tweet2vec}
Vosoughi, S., Vijayaraghavan, P., Roy, D.: Tweet2vec: Learning tweet embeddings
  using character-level cnn-lstm encoder-decoder. In: Proceedings of the 39th
  International ACM SIGIR conference on Research and Development in Information
  Retrieval. pp. 1041--1044 (2016)

\bibitem{wu2021exploring}
Wu, H., Zhang, Z., Wu, Q.: Exploring syntactic and semantic features for
  authorship attribution. Applied Soft Computing  \textbf{111},  107815 (2021)

\bibitem{zhang2018syntax}
Zhang, R., Hu, Z., Guo, H., Mao, Y.: Syntax encoding with application in
  authorship attribution. In: Proceedings of the 2018 Conference on Empirical
  Methods in Natural Language Processing. pp. 2742--2753 (2018)

\bibitem{zheng2023review}
Zheng, W., Jin, M.: A review on authorship attribution in text mining. Wiley
  Interdisciplinary Reviews: Computational Statistics  \textbf{15}(2),  e1584
  (2023)

\end{thebibliography}
\end{document}